\pdfoutput=1

\documentclass[11pt]{article}

\usepackage{acl}

\usepackage{times}
\usepackage{latexsym}

\usepackage[T1]{fontenc}

\usepackage[utf8]{inputenc}

\usepackage{microtype}

\usepackage{microtype}
\usepackage{graphics}
\usepackage{graphicx}
\usepackage{caption}
\usepackage{subcaption}
\usepackage{multirow}
\usepackage{hhline}
\usepackage{adjustbox}
\usepackage{mathtools}
\usepackage{amsfonts}
\usepackage{tabularx}
\usepackage{enumitem}
\usepackage{booktabs}

\title{Towards Structure-aware Paraphrase Identification with Phrase Alignment Using Sentence Encoders}

\author{Qiwei Peng~~~David Weir~~~Julie Weeds\\
  University of Sussex \\
  Brighton, UK \\
  \texttt{\{qiwei.peng, d.j.weir, j.e.weeds\}@sussex.ac.uk} \\}

\begin{document}
\maketitle
\begin{abstract}
Previous works have demonstrated the effectiveness of utilising pre-trained sentence encoders based on their sentence representations for meaning comparison tasks. Though such representations are shown to capture hidden syntax structures, the direct similarity comparison between them exhibits weak sensitivity to word order and structural differences in given sentences. A single similarity score further makes the comparison process hard to interpret. Therefore, we here propose to combine sentence encoders with an alignment component by representing each sentence as a list of predicate-argument spans (where their span representations are derived from sentence encoders), and decomposing the sentence-level meaning comparison into the alignment between their spans for paraphrase identification tasks. Empirical results show that the alignment component brings in both improved performance and interpretability for various sentence encoders. After closer investigation, the proposed approach indicates increased sensitivity to structural difference and enhanced ability to distinguish non-paraphrases with high lexical overlap.
\end{abstract}








\section{Introduction}
\label{introduction}
Sentence meaning comparison measures the semantic similarity of two sentences. Specifically, the task of paraphrase identification binarises the similarity as paraphrase or non-paraphrase depending on whether they express similar meanings \citep{bhagat2013paraphrase}. This task benefits many natural language understanding applications, like plagiarism identification \citep{chitra2016plagiarism} and fact checking \citep{10.1145/3366423.3380231}, where it is important to detect same things said in different ways.

\begin{table*}[]
\Large
\centering
\renewcommand{\arraystretch}{1.4}
\resizebox{\textwidth}{!}{%
\begin{tabular}{|c|l|l|c|}
\hline
\multicolumn{1}{|l|}{Dataset} & Sentence A                                                                                                   & Sentence B                                                                                                                 & Label \\ \hline
\multirow{2}{*}{MSRP}         & \begin{tabular}[c]{@{}l@{}}The Toronto Stock Exchange opened on time and \\ slightly lower.\end{tabular}     & \begin{tabular}[c]{@{}l@{}}The Toronto Stock Exchange said it will be business \\ as usual on Friday morning.\end{tabular} & N     \\ \cline{2-4} 
                              & \begin{tabular}[c]{@{}l@{}}More than half of the songs were purchased as \\ albums, Apple said.\end{tabular} & \begin{tabular}[c]{@{}l@{}}Apple noted that half the songs were purchased \\ as part of albums.\end{tabular}               & Y     \\ \hline
\multirow{2}{*}{PAWS}         & What factors cause a good person to become bad?                                                              & What factors cause a bad person to become good?                                                                            & N     \\ \cline{2-4} 
                              & The team also toured in Australia in 1953.                                                                    & In 1953, the team also toured in Australia.                                                                                & Y     \\ \hline
\end{tabular}%
}
\caption{Example sentence pairs taken from both MSRP and PAWS datasets. Y stands for paraphrases while N stands for non-paraphrases.}
\label{tab:dataset-example}
\end{table*}

The difference in sentence structures is important for distinguishing their meanings. However, as shown in Table \ref{tab:dataset-example} and \ref{tab:lexical-overlap}, many existing paraphrase identification datasets exhibit high correlation between positive pairs and the degree of their lexical overlap, such as the Microsoft Research Paraphrase Corpus (MSRP) \citep{dolan-brockett-2005-automatically}. Models trained on them tend to mark sentence pairs with high word overlap as paraphrases despite clear clashes in meaning. In light of this, \citet{zhang2019paws} utilised word scrambling and back translation to create the Paraphrase Adversaries from Word Scrambling (PAWS) datasets which are mainly concerned with word order and structure by creating paraphrase and non-paraphrase pairs with high lexical overlap. As also shown in these two tables, sentence pairs in the PAWS datasets demonstrate much higher lexical overlap and lower correlation, which requires models to pay more attention to word order and sentence structure to successfully distinguish non-paraphrases from paraphrases.

Recently, various pre-trained sentence encoders have been proposed to produce high-quality sentence embeddings for downstream usages \cite{reimers2019sentence, thakur-etal-2021-augmented, gao2021simcse}. Such embeddings are compared to derive a similarity score for different meaning comparison tasks, including paraphrase identification. Though widely used, sentence encoders still face challenges from different aspects in case of meaning comparison. Pre-trained models are observed to capture structural information to some extent \citep{clark2019does,hewitt2019structural,jawahar2019does}. However, as we will demonstrate in this work, their direct comparison of two sentence vectors performs poorly on PAWS datasets indicating weak sensitivity to structural difference, though they achieve good performance on other general paraphrase identification datasets like MSRP. In addition, the single similarity score derived from the comparison of two vectors is difficult to interpret. This thus motivates us to find a better way of utilising sentence encoders for meaning comparison.
 
Elsewhere, researchers have worked on decomposing sentence-level meaning comparison into comparisons at a lower level, such as word and phrase-level, which largely increased the interpretability \citep{he2016pairwise, chen2017enhanced, zhang2019explicit}. Alignment is the core component in these proposed systems, where sentence units at different levels are aligned through either training signals or external linguistic clues, after which a matching score is derived for sentence-level comparison. Here, we argue that, instead of comparing sentence meaning by using sentence embeddings, it would be better to combine sentence encoders with alignment components in a structure-aware way to strengthen the sensitivity to structural difference and to gain interpretability.

An important aspect of sentence meaning is its predicate-argument structure, which has been utilised in machine translation \citep{xiong2012modeling} and paraphrase generation \citep{ganitkevitch2013ppdb, kozlowski2003generation}. Given the importance of detecting structural differences in paraphrase identification tasks, we propose to represent each sentence as a list of predicate-argument spans where span representations are derived from sentence encoders, and to decompose sentence-level meaning comparison into the direct comparison between their aligned predicate-argument spans by taking advantage of the Hungarian algorithm \citep{kuhn1956variants, 7738348}. The sentence-level score is then derived by aggregation over their aligned spans. Without re-training, the proposed alignment-based sentence encoder can be used with enhanced structure-awareness and interpretability.

As pre-trained sentence encoders produce contextualised representations, two phrases of different meaning might be aligned together due to their similar syntactic structure and contexts. For example: 
\begin{itemize}
    \item[a)] \textit{Harris announced on twitter that he will quit.}
    \item[b)] \textit{James announced on twitter that he will quit.}
\end{itemize}
Unsurprisingly, the span \textit{Harris announced} will be aligned to the span \textit{James announced} with a high similarity score given that they share exactly the same context and syntactic structure. However, it might be problematic to consider this high similarity score when we calculate the overall score given clear clashes in the meaning at sentence-level. In this regard, we further explore how the contextualisation affects paraphrase identification by comparing aligned phrases based on their de-contextualised representations.

Empirical results show that the inclusion of the alignment component leads to improvements on four paraphrase identification tasks and demonstrates increased ability to detect non-paraphrases with high lexical overlap, plus an enhanced sensitivity to structural difference. Upon closer investigation, we find that applying de-contextualisation to aligned phrases could further help to recognise such non-paraphrases. 

In summary, our contributions are as follows:
\begin{itemize}[noitemsep,topsep=0.3pt]
    \item[1)] We propose an approach that combines sentence encoders with an alignment component by representing sentences as lists of predicate-argument spans and decomposing sentence-level meaning comparison into predicate-argument span comparison. 
    \item[2)] We provide an evaluation on four different paraphrase identification tasks, which demonstrates both the improved sensitivity to structures and the interpretability at inference time.
    \item[3)] We further introduce a de-contextualisation step which can benefit tasks that aim to identify non-paraphrases of extremely high lexical overlap.
\end{itemize}

\section{Related Work}
\subsection{Sentence Encoders}
Sentence encoders have been studied extensively in years. \citet{kiros2015skip} abstracted the skip-gram model \citep{DBLP:journals/corr/abs-1301-3781} to the sentence level and proposed Skip-Thoughts by using a sentence to predict its surrounding sentences in an unsupervised manner. InferSent \citep{conneau2017supervised}, on the other hand, leveraged supervised learning to train a general-purpose sentence encoder with BiLSTM by taking advantage of natural language inference (NLI) datasets. Pre-trained language models like BERT \citep{devlin-etal-2019-bert} are widely used to provide a single-vector representation for the given sentence and demonstrate promising results across a variety of NLP tasks. Inspired by InferSent, Sentence-BERT (SBERT) \citep{reimers2019sentence} produces general-purpose sentence embeddings by fine-tuning BERT on NLI datasets. However, as investigated by \citet{li-etal-2020-sentence}, sentence embeddings produced by pre-trained models suffer from anisotropy, which severely limits their expressiveness. They then proposed a post-processing step to map sentence embeddings to an isotropic distribution which largely improves the situation. Similarly, \citet{su2021whitening} proposed a whitening operation for post-process, which aims to alleviate the anisotropy problem. \citet{gao2021simcse}, on the other hand, proposed the SimCSE model by fine-tuning pre-trained sentence encoders with a contrastive learning objective \citep{chen2020simple} along in-batch negatives \citep{henderson2017efficient, chen2017enhanced} on NLI datasets, improving both the performance and the anisotropy problem. Though sentence encoders have achieved promising performance, the current way of utilising them for meaning comparison tasks has known drawbacks and could benefit from the fruitful developments of the alignment component, which have been widely used in modelling sentence pair relations.

\subsection{Alignment in Sentence Pair Tasks}
Researchers have been investigating sentence meaning comparison for years. One widely used method involves decomposing the sentence-level comparison into comparisons at a lower level. \citet{maccartney-etal-2008-phrase} aligned phrases based on their edit distance and applied the alignment to NLI tasks by taking average of aligned scores. \citet{5360085} decomposed sentence-level similarity score into the direct comparison between events and content words based on WordNet \citep{miller1995wordnet}. \citet{sultan2014dls} proposed a complex alignment pipeline based on various linguistic features, and predicted the sentence-level semantic similarity by taking the proportion of their aligned content words. The alignment between two syntactic trees are used along with other lexical and syntactic features to determine whether two sentences are paraphrases with SVM \citep{liang2016learning}.

Similar ideas are combined with neural models to construct alignments based on the attention mechanism \citep{DBLP:journals/corr/BahdanauCB14}. They can be seen as learning soft alignments between words or phrases in two sentences. \citet{pang2016text} proposed MatchPyramid where a word-level alignment matrix was learned, and convolutional networks were used to extract features for sentence-level classification. More fine-grained comparisons between words are introduced by PMWI \citep{he2016pairwise} to better dissect the meaning difference. \citet{wang2016sentence} put focus on both similar and dissimilar alignments by decomposing and composing lexical semantics over sentences. ESIM \citep{chen2017enhanced} further allowed richer interactions between tokens. These models are further improved by incorporating context and structure information \citep{liu2019incorporating}, as well as character-level information \citep{lan2018character}. Recently, Pre-trained models are exploited to provide contextualised representations for the PMWI \citep{zhang2019explicit}. Instead of relying on soft alignments, some other models tried to take the phrase alignment task as an auxiliary task for sentence semantic assessments \citep{arase-tsujii-2019-transfer,ARASE2021101164}, and to embed the Hungarian algorithm into trainable end-to-end neural networks to provide better aligned parts \citep{XIAO2020172}. Considering pre-trained sentence encoders are often directly used to provide fixed embeddings for meaning comparison, in this work, we propose to combine them with the alignment component at inference time so that it can be used with enhanced structure-awareness without re-training.

\begin{figure*}[]
\centering
\includegraphics[width=1.9\columnwidth]{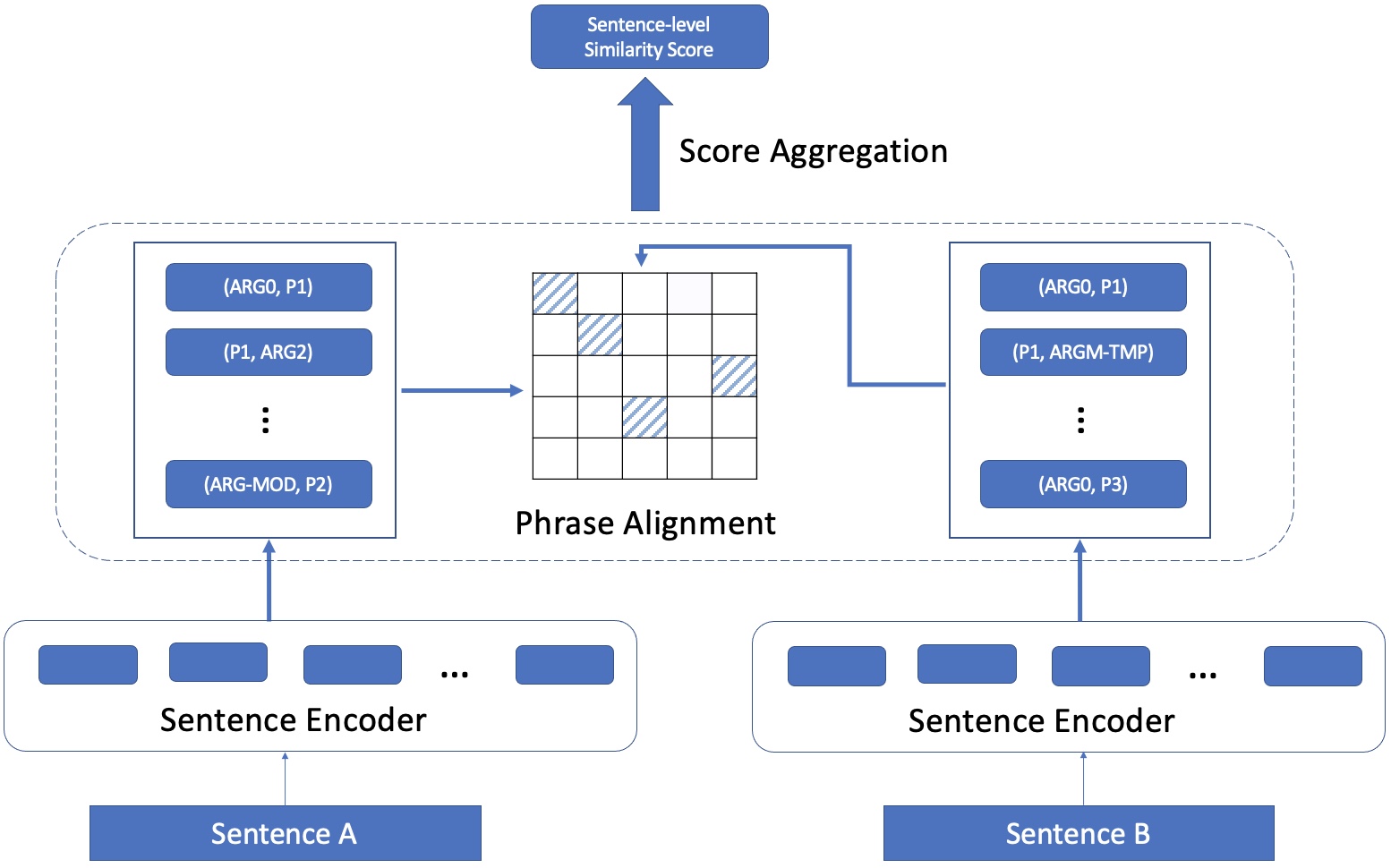}
\caption{The proposed approach for paraphrase identification that combines sentence encoders with the phrase alignment at inference time. Predicate-argument spans are first extracted from sentences. Span representations are then derived from contextualised token representations. We perform Hungarian algorithm to align extracted phrase spans and obtain the sentence-level similarity score by aggregation over aligned spans. The alignment matrix is useful for interpretation.}
\label{fig:model_structure}
\end{figure*}  

\section{Our Approach}
Instead of generating a single-vector representation for meaning comparison based on sentence encoders, we propose to represent each sentence as a list of predicate-argument spans and use sentence encoders to provide its span representations. The comparison between two sentences is then based on the alignment between their predicate-argument spans. As depicted in Figure \ref{fig:model_structure}, the approach can be considered as a post-processing step and consists of the following main components:

\paragraph{Sentence Encoders:} The input sentences are first fed into sentence encoders to produce contextualised token representations that will later be used to create context-aware phrase representations from the last hidden layer. The phrase representation will be the basic unit of our meaning comparison method.  

\begin{figure}[htb!]
\centering
\includegraphics[width=\columnwidth]{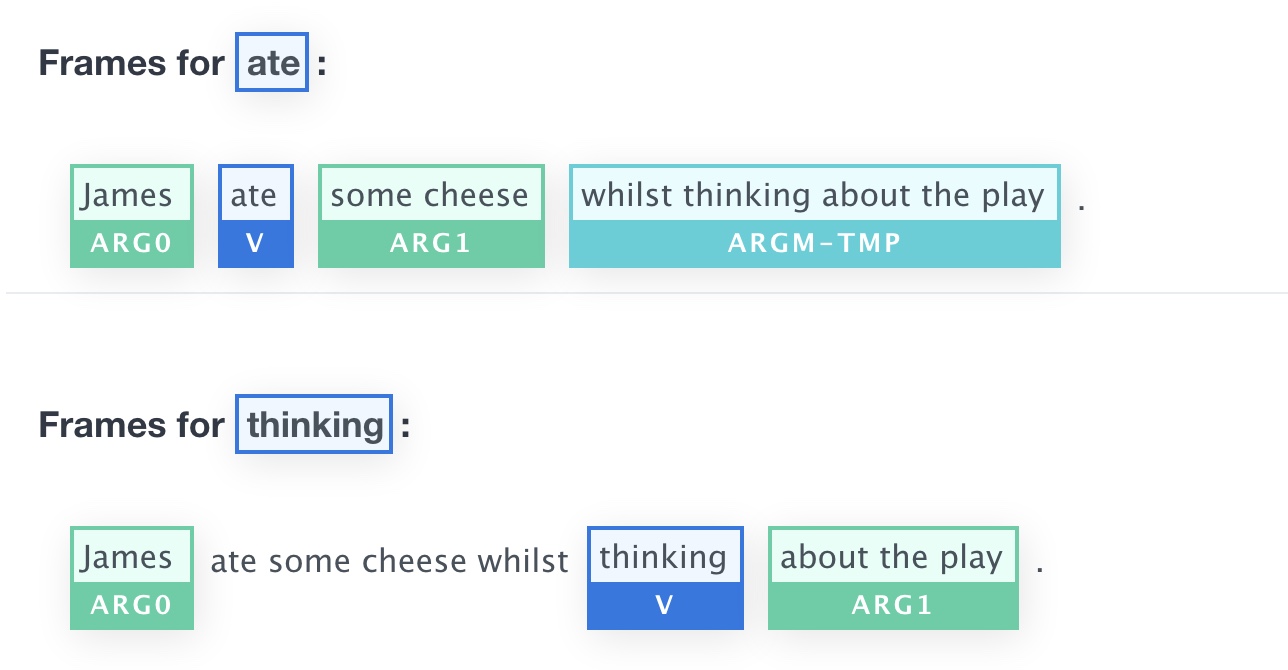}
\caption{The extracted predicates and relevant semantic arguments for the given example sentence. Outputs are produced by the AllenNLP SRL tagger.}
\label{fig:srl_tagger}
\end{figure}

\paragraph{Predicate Argument Spans (PAS):} For each sentence, we first apply a BERT-based semantic role labelling (SRL) tagger provided by AllenNLP \citep{gardner2018allennlp} to obtain both predicates and relevant arguments for each sentence. To generate predicate argument spans, we group the predicate and its arguments together and order them according to their original position in the sentence. Following is an example of predicate-argument spans from a sentence:

\begin{quote} 
\textit{James ate some cheese whilst thinking about the play.}
\end{quote}
Two predicates, \textit{ate} and \textit{thinking}, are extracted by the tagger. As shown in Figure \ref{fig:srl_tagger}, a number of arguments with different relations are discovered for each predicate. We further group them into predicate-argument spans. For the given sentence, we will have three spans for the predicate \textit{ate}: (\textit{James}, \textit{ate}), (\textit{ate}, \textit{some}, \textit{cheese}), (\textit{ate}, \textit{whilst}, \textit{thinking}, \textit{about}, \textit{the}, \textit{play}) and two spans for the predicate \textit{thinking}: (\textit{James}, \textit{thinking}), (\textit{thinking}, \textit{about}, \textit{the}, \textit{play}). If no predicate or associated argument is found, we take the whole sentence itself as a long span.

\begin{figure}[htb!]
\centering
\includegraphics[width=\columnwidth]{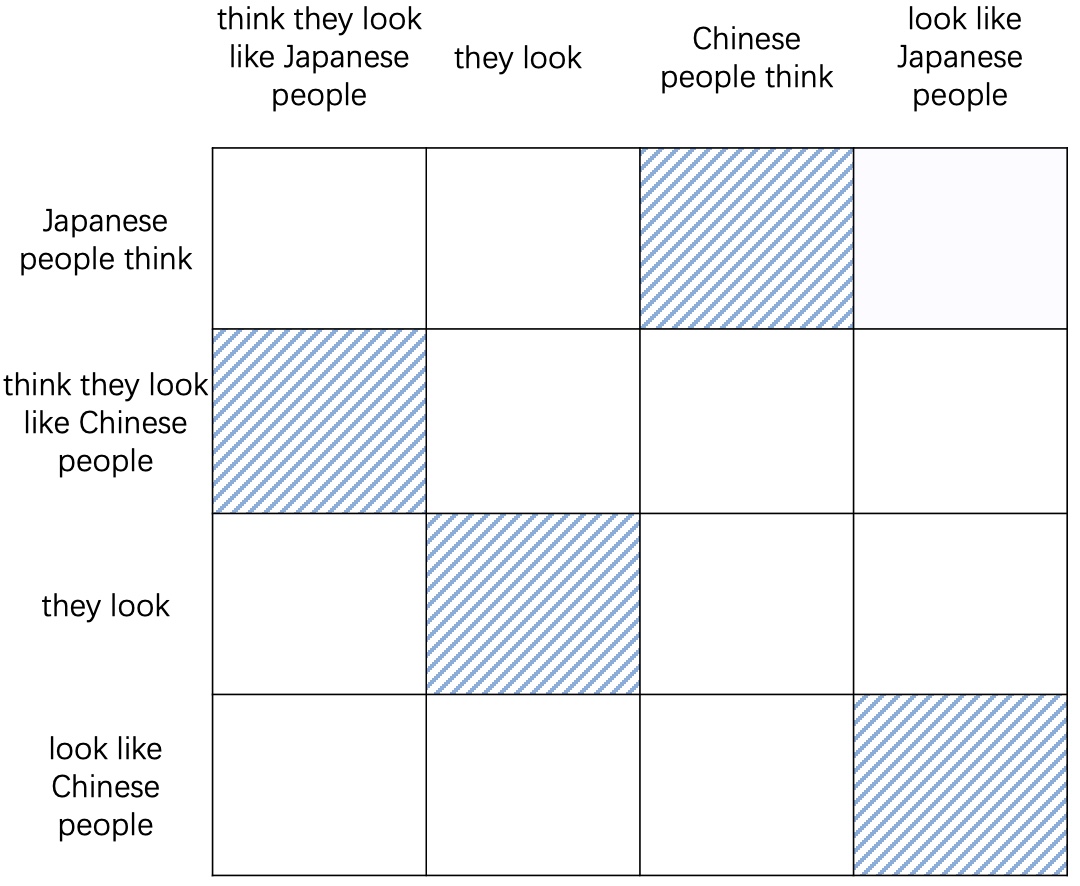}
\caption{The predicate-argument span alignment between the example pair taken from PAWS\_QQP, sentence A: \textit{ do Chinese people think they look like Japanese people?} and sentence B: \textit{ do Japanese people think they look like Chinese people?}}
\label{fig:alignment-example}
\end{figure}  

\paragraph{Phrase Alignment:} After obtaining all predicate-argument spans, we derive their span representations based on the used encoder. As previous works have shown, aligning with contextual information could achieve better performance and help with disambiguation \citep{arase2020compositional,dou2021word}. We take the mean-pooling over all tokens in the span to produce a contextualised span representation for later alignment. The tokenization strategy in BERT generates sub-tokens, whereas in the produced spans, we have word tokens. To align them properly, we use the same tokenizer to break the original word into sub-tokens and represent it as a list of sub-tokens in the span if a sub-token exists. Given two collections of predicate-argument span representations, $p = \{p_1, p_2, ..., p_M\}$ and $q = \{q_1, q_2, ..., q_N\}$, we are trying to find the best alignments between them. This can be viewed as a standard assignment problem that has been extensively handled by Hungarian algorithm \citep{kuhn1956variants}. A similarity matrix, $C$, is constructed for each pair of sentences where the row has one collection of spans and the column has another. The value for each entry cell, $C_{mn}$, is the cosine similarity score between the two span, $p_m$ and $q_n$. The task of finding the best alignment is to find alignments among two collections that give the maximum score:

\begin{equation}
max\sum_{m}\sum_{n}C_{mn}X_{mn}
\end{equation}
$X$ is a boolean matrix where $X[m,n]=1$ if span $m$ is assigned to span $n$. We apply the modified Jonker-Volgenant algorithm\footnote{We use its Scipy implementation: \url{https://docs.scipy.org/doc/scipy/reference/generated/scipy.optimize.linear\_sum\_assignment.html}} \citep{7738348} to find the best alignments that maximise the overall score. After discovering the optimal $X$, we obtain a collection of aligned span pairs associated with their cosine similarity scores, $A = \{A_1, A_2, ..., A_l\}$. One alignment example taken from PAWS is given in Figure \ref{fig:alignment-example}.

\paragraph{Score Aggregation:} To produce a sentence-level similarity score for the given pair, we simply take the mean-pooling over scores of all aligned parts:
\begin{equation}
Score_{ij} = MeanPooling(A_1,..,A_l)
\end{equation}
The similarity score between sentence $i$ and sentence $j$ is the average score of their aligned spans, and will be used for determining whether the sentence pair is paraphrase or non-paraphrase. The alignment matrix, as shown in Figure \ref{fig:alignment-example}, is useful to explain how the overall score is derived and why.

\section{Experiments}
We follow the same two-step procedure in previous work for evaluation \citep{li-etal-2020-sentence, thakur-etal-2021-augmented}. For vanilla sentence encoders, we first generate fixed sentence embeddings, and then derive sentence-level similarity scores by calculating the cosine similarity between two embeddings. For sentence encoders combined with the alignment component, we derive sentence-level similarity scores by aggregation over cosine scores of all aligned spans where span representations are derived from sentence encoders. Otherwise specifically stated, the alignment is performed between predicate-argument spans (PAS). Their performances under these two scenarios are evaluated and compared. We here experiment with three widely used sentence encoders, BERT-base \citep{devlin-etal-2019-bert}, Sentence-BERT (SBERT) \citep{reimers2019sentence} and SimCSE \citep{gao2021simcse}.

\begin{table}[h!]
\centering
\renewcommand{\arraystretch}{1.2} 
\begin{tabular}{|c|c|c|c|}
\hline
Datasets   & Train  & Dev   & Test  \\ \hline
PAWS\_QQP  & 11,986  & -  & 677   \\ \hline
PAWS\_Wiki & 49,401  & 8,000  & 8,000  \\ \hline
MSRP       & 3,668   & 408   & 1,725  \\ \hline
TwitterURL & 37,976  & 4,224  & 9,334  \\ \hline
\end{tabular}
\caption{Statistics of all four datasets used in this work.}
\label{tab:data-statistics}
\end{table}

\subsection{Datasets}
In this work, we evaluate the proposed approach on four different paraphrase identification tasks. The statistics of these datasets are listed in Table \ref{tab:data-statistics}. Below we give some basic descriptions: 
\begin{itemize}
  \item \textbf{PAWS\_QQP}: In order to assess the sensitivity to word order and syntactic structure, \citet{zhang2019paws} proposed a paraphrase identification dataset which has extremely high lexical overlap by applying back translation and word scrambling to sentences taken from the Quora Question Pairs \citep{wang2017bilateral}.
  \item \textbf{PAWS\_Wiki}: Similar to PAWS\_QQP, the same technique is applied to sentences obtained from Wikipedia articles to construct sentence pairs \citep{zhang2019paws}. Both PAWS datasets aim to measure sensitivity of models on word order and sentence structure.
  \item \textbf{Microsoft Research Paraphrase Corpus (MSRP)}: This corpus constructs sentence pairs by clustering news articles with an SVM classifier and human annotations \citep{dolan-brockett-2005-automatically}. It has 4,076 train data and 1,725 test data. In this paper, we adopt the same split strategy as stated in GLUE \citep{DBLP:conf/iclr/WangSMHLB19}.
  \item \textbf{TwitterURL}: \citet{lan2017continuously} proposed the TwitterURL corpus where sentence pairs in the dataset are collected by linking tweets that share the same URL of news articles. The corpus contains multiple references of both formal well-edited and informal user-generated texts.
\end{itemize}

\begin{table}[h!]
\centering
\renewcommand{\arraystretch}{1.2}
\resizebox{0.85\columnwidth}{!}{%
\begin{tabular}{|c|ccc|}
\hline
\multicolumn{1}{|c|}{\multirow{2}{*}{Datasets}} & \multicolumn{3}{c|}{Lexical Overlap}                                    \\ \cline{2-4} 
\multicolumn{1}{|c|}{}                          & \multicolumn{1}{c|}{Positive} & \multicolumn{1}{c|}{Negative} & Overall \\ \hline
PAWS\_QQP                                       & \multicolumn{1}{c|}{95.24\%}  & \multicolumn{1}{c|}{96.79\%}  & 96.35\% \\ \hline
PAWS\_Wiki                                      & \multicolumn{1}{c|}{84.50\%}  & \multicolumn{1}{c|}{84.99\%}  & 84.77\% \\ \hline
MSRP                                            & \multicolumn{1}{c|}{55.95\%}  & \multicolumn{1}{c|}{42.60\%}  & 51.48\% \\ \hline
TwitterURL                                      & \multicolumn{1}{c|}{29.28\%}  & \multicolumn{1}{c|}{7.73\%}   & 11.94\% \\ \hline
\end{tabular}%
}
\caption{The lexical overlap between sentence pairs across different datasets. We report both the overall figure and the figures for each class. We calculate the lexical overlap in terms of Jaccard Similarity with ngram=1.}
\label{tab:lexical-overlap}
\end{table}

The percentage of lexical overlap between sentence pairs in terms of their labels are summarised in Table \ref{tab:lexical-overlap}. It shows that sentence pairs taken from the PAWS datasets generally have higher lexical overlap. Compared to datasets like MSRP and TwitterURL, where negative examples have lower lexical overlap than positive examples, the two PAWS datasets exhibit similar degrees of lexical overlap regardless of their labels. In light of this, we expect that models that are sensitive to word order and sentence structure would demonstrate greater improvements on the PAWS datasets in comparison to models without such sensitivity. Specifically, we put our focus on the PAWS datasets and explore whether different models capture structural differences.

\subsection{Implementation Details}
For sentence encoders used in this work, we generate sentence embeddings according to their default strategies. For BERT-base\footnote{We use its huggingface implementation: \url{https://huggingface.co/bert-base-uncased}} and SBERT\footnote{\url{https://github.com/UKPLab/sentence-transformers}}, we use the mean-pooling over the last hidden layer as its sentence representation, and for SimCSE\footnote{\url{https://github.com/princeton-nlp/SimCSE}}, we use the CLS token after the trained MLP layer. For all experiments in this work, no training process is involved. In order to calculate evaluation metrics like accuracy and F1 score, we find optimal thresholds for different metrics on the development set, and apply them on test sets to binary the cosine similarity as paraphrase or non-paraphrase. Given PAWS\_QQP does not have development set, we randomly sample 20\% of its training data as the development set following the same class distribution. All experiments are conducted on RTX 3090 GPUs.

\begin{table*}[h]
\centering
\renewcommand{\arraystretch}{1.2}
\resizebox{0.8\textwidth}{!}{%
\begin{tabular}{|c|c|c|c|c|c|}
\hline
                             & \begin{tabular}[c]{@{}c@{}}PAWS\_QQP\\ (F1/ACC)\end{tabular}                                 & \begin{tabular}[c]{@{}c@{}}PAWS\_Wiki\\ (F1/ACC)\end{tabular}                                & \begin{tabular}[c]{@{}c@{}}TwitterURL\\ (F1/ACC)\end{tabular}                                & \begin{tabular}[c]{@{}c@{}}MSRP\\ (F1/ACC)\end{tabular}                                      & \begin{tabular}[c]{@{}c@{}}AVG\\ (F1/ACC)\end{tabular}                  \\ \hline
BERT                         & 37.13/72.97                               & 61.28/56.75                               & 63.24/85.65                               & 80.50/70.38                               & 60.54/71.44          \\
+ Alignment                        & \textbf{47.46/75.18}                      & \textbf{63.08/62.58}                      & \textbf{65.26/86.52}                      & \textbf{80.96/70.61}                      & \textbf{64.19/73.72} \\ \hline
SBERT                        & 33.95/74.74                               & 61.83/60.63                               & 65.61/87.04                               & 81.68/73.39                               & 60.61/73.95          \\
+ Alignment                        & \textbf{52.75/77.70}                      & \textbf{62.52/64.51}                      & \textbf{66.60/87.33}                      & \textbf{82.10/73.80}                      & \textbf{65.99/75.84} \\ \hline
\multicolumn{1}{|c|}{SimCSE} & \multicolumn{1}{l|}{36.16/75.48}          & \multicolumn{1}{c|}{61.32/62.58}          & \multicolumn{1}{l|}{\textbf{69.20/87.74}} & \multicolumn{1}{c|}{\textbf{82.80/74.61}} & 62.37/75.10          \\
\multicolumn{1}{|c|}{+ Alignment}  & \multicolumn{1}{l|}{\textbf{57.49/79.17}} & \multicolumn{1}{c|}{\textbf{65.00/65.99}} & \multicolumn{1}{l|}{67.83/87.27}          & \multicolumn{1}{c|}{81.70/73.68}          & \textbf{68.01/76.53} \\ \hline
\end{tabular}%
}
\caption{Results on four paraphrase identification tasks, we report both the F1 score of the positive class and the overall accuracy. Cells marked bold have the best performance in each column.}
\label{tab:main-table}
\end{table*}

\subsection{Evaluation}

The main results are summarised in Table \ref{tab:main-table}, and we report the F1 score of the positive class as well as the overall accuracy. It shows that, with our proposed approach, the performance of different sentence encoders can generally be improved. In addition, significant improvements are observed on PAWS datasets after we introduce the alignment component. This demonstrates the effectiveness of our proposed alignment-based sentence encoder and validates the improved sensitivity to word order and sentence structure. Furthermore, we find that the performance of different models, regardless of combining with the alignment component, is similar or competitive on MSRP and TwitterURL datasets. It suggests that both datasets are inadequate when used to detect the model's structure-awareness for the structural information is not required to achieve high scores on them. Accordingly, compared to its alignment version, the lack of sensitivity to structural differences translates the slightly better performance on TwitterURL and MSRP obtained by SimCSE into much worse performance on the PAWS datasets. This further supports our previous arguments and demonstrates the advantages of introducing the alignment component to enhance structure-awareness.

\section{Analysis}
To better understand the improvements, we have conducted several experiments to investigate different aspects of the proposed approach. Given we are mostly interested in the performance on the two PAWS datasets, we only experiment and report the results on the PAWS\_QQP and PAWS\_Wiki in the following experiments.

\subsection{Comparison to Other Span Strategies}

\begin{table}[]
\centering
\renewcommand{\arraystretch}{1.0}
\resizebox{\columnwidth}{!}{%
\begin{tabular}{|l|c|c|}
\hline
Models                  & \begin{tabular}[c]{@{}c@{}}PAWS\_QQP\\ (F1/ACC)\end{tabular} & \begin{tabular}[c]{@{}c@{}}PAWS\_Wiki\\ (F1/ACC)\end{tabular} \\ \hline
BERT-TokenLevel         & 40.13/73.41                                                  & 62.33/62.36                                                   \\
BERT-RandomSpan         & 19.91/73.71                                                  & 61.30/57.19                                                   \\
BERT-ContinuousRandom   & 39.86/74.74                                                  & 61.25/57.66                                                   \\
BERT-PAS                & \textbf{47.46/75.18}                                         & \textbf{63.08/62.58}                                          \\ \hline
SBERT-TokenLevel        & 47.51/75.04                                                  & 61.65/64.15                                                   \\
SBERT-RandomSpan        & 38.89/73.12                                                  & 61.07/59.08                                                   \\
SBERT-ContinuousRandom  & 46.56/74.74                                                  & 61.28/58.49                                                   \\
SBERT-PAS               & \textbf{52.75/77.70}                                         & \textbf{62.52/64.51}                                          \\ \hline
SimCSE-TokenLevel       & 50.74/74.00                                                  & 62.03/63.28                                                   \\
SimCSE-RandomSpan       & 34.31/73.56                                                  & 61.24/57.70                                                   \\
SimCSE-ContinuousRandom & 40.74/77.25                                                  & 61.30/57.24                                                   \\
SimCSE-PAS              & \textbf{57.49/79.17}                                         & \textbf{65.00/65.99}                                          \\ \hline
\end{tabular}%
}
\caption{Evaluation using different span types for alignment. We report the F1 score of the positive class and the overall accuracy.}
\label{tab:span-type}
\end{table}

In this experiment, we consider three more scenarios with different span types, and investigate the impact of the predicate-argument span. The alignment between different tokens are widely used in previous works, so here, instead of aligning predicate-argument spans, we directly conduct alignment at token-level. Two further strategies are explored regarding phrase-level alignment. Firstly, we randomly sample words from the sentence to make a span, where the words in each span might not necessarily be sequential. In the RandomSpan strategy, no linguistically-meaningful structures are preserved. Secondly, we randomly sample continuous word sequences to build a span, where the span must contain sequential texts. In this ContinuousRandom strategy, only sequential relations are preserved. The length of the sampled spans is arbitrary. To make a fair comparison, the number of sampled spans is the same as that of the predicate-argument spans in the sentence. As demonstrated in Table \ref{tab:span-type}, the alignment between predicate-argument spans outperforms all the others. In other words, the model's sensitivity to word order and structural differences can be greatly improved by comparing two sentences' predicate-argument structures. 
\subsection{Large Improvements in Recall}

\begin{table}[]
\centering
\resizebox{0.8\columnwidth}{!}{%
\begin{tabular}{|c|c|c|}
\hline
      & \begin{tabular}[c]{@{}c@{}}PAWS\_QQP\\ (recall of +)\end{tabular} & \begin{tabular}[c]{@{}c@{}}PAWS\_Wiki\\ (recall of -)\end{tabular} \\ \hline
BERT   & 32.46                                                             & 0.09                                                               \\
+ Alignment  & \textbf{36.65}                                                    & \textbf{25.96}                                                     \\ \hline
SBERT  & 24.08                                                             & 9.14                                                               \\
+ Alignment  & \textbf{47.64}                                                    & \textbf{29.53}                                                     \\ \hline
SimCSE & 25.65                                                             & 0.27                                                               \\
+ Alignment  & \textbf{50.26}                                                    & \textbf{52.28}                                                     \\ \hline
\end{tabular}%
}
\caption{Results on PAWS\_QQP and PAWS\_Wiki. For PAWS\_QQP, we report the recall of positive class and for PAWS\_Wiki, we report the recall of negative class.}
\label{tab:paws-recall}
\end{table}

We have observed significant improvements on PAWS datasets by introducing the alignment between predicate-argument spans in previous experiments. It is crucially important to understand how the improvement is obtained. In this experiment, we look into the recall of positive and negative pairs. In PAWS\_Wiki, we find that almost all sentence pairs are classified as positive by vanilla models given the near-zero recall for the negative class as shown in Table \ref{tab:paws-recall}. Despite utterly incorrect predictions, it spuriously lowers the performance gap (on PAWS\_Wiki) in terms of the F1 score of the positive class as shown in Table \ref{tab:main-table}. After applying the alignment process to sentence encoders, we notice significant improvements in the recall of negative class. About 70\% of sentence pairs in the PAWS\_QQP have negative labels, which makes vanilla models difficult to distinguish paraphrases from non-paraphrases and mark most of sentence pairs as negative, as evidenced by the low recall for positives in the table. Similarly, we observe significant improvements in recall after introducing the alignment component. The large improvements in recall demonstrate the enhanced ability to distinguish non-paraphrases from paraphrases. Moreover, as shown in Table \ref{tab:main-table}, the improvements in recall are not at the expense of their general performance, since we are still improving on F1 scores and the overall accuracy.

\subsection{De-contextualisation}

\begin{table}[]
\centering
\renewcommand{\arraystretch}{1.1}
\resizebox{\columnwidth}{!}{%
\begin{tabular}{|l|l|l|}
\hline
Models      & \begin{tabular}[c]{@{}l@{}}PAWS\_QQP\\ (F1/recall of + )\end{tabular} & \begin{tabular}[c]{@{}l@{}}PAWS\_Wiki\\ (F1/recall of - )\end{tabular} \\ \hline
BERT-Alignment    & 47.46/36.65                                                           & 63.08/25.96                                                            \\
+ decontext & \textbf{52.50/43.98}                                                  & \textbf{63.39/45.09}                                                   \\ \hline
SBERT-Alignment   & 52.75/47.64                                                           & 62.52/29.53                                                            \\
+ decontext & \textbf{65.43/64.40}                                                  & \textbf{66.63/64.38}                                                   \\ \hline
SimCSE-Alignment  & 57.49/50.26                                                           & 65.00/52.28                                                            \\
+ decontext & \textbf{65.16/68.06}                                                  & \textbf{67.32/54.14}                                                   \\ \hline
\end{tabular}%
}
\caption{The results on PAWS datasets after applying de-contextualisation. We report the F1 score of the positive class on both datasets, the recall of positive class on PAWS\_QQP, and the recall of negative class on PAWS\_Wiki.}
\label{tab:decontext}
\end{table}

As pre-trained sentence encoders produce contextualised representations, two phrases of different meaning might be aligned for similar syntactic structure and contexts with a high similarity score. In the example shown in Section \ref{introduction}, \textit{Harris announced} will be aligned with \textit{James announced} with a high similarity score given their identical syntactic structure and contexts. However, does such high similarity score make sense when it comes to the task of paraphrase identification? Comparing the meaning of two phrases in the context of their use often helps disambiguate. In this case, the highly similar context instead downplays the difference, while it is the minor difference that changes the whole sentence meaning. This problem is exacerbated in the PAWS datasets given that both PAWS\_QQP and PAWS\_Wiki have extremely high lexical overlap, with 96.35\% and 84.77\% respectively, as shown in Table \ref{tab:lexical-overlap}. Such high lexical overlap indicates a similar context, and thus a high similarity score between aligned phrases. In this experiment, we align phrases based on their contextualised representations as before but de-contextualise them by sending these phrases, without context, through sentence encoders to produce context-agnostic representations. A similarity score at sentence-level is then derived from cosine similarities between context-agnostic representations. 

We show the results in Table \ref{tab:decontext}. It clearly shows that, in spite of losing contextual information, the model with de-contextualisation process appears to improve the performance significantly. Additionally, it suggests that contextualisation might be harmful in situations where we focus on small differences that might change the meaning of the whole.

\section{Conclusion}
In this work, we propose an approach that combines sentence encoders with an alignment component by representing sentences as lists of predicate-argument spans and decomposing sentence-level meaning comparison into predicate-argument span comparison. Experiments with three widely used sentence encoders show that such method leads to improvements on various paraphrase identification tasks and increases the sensitivity to word order and structural differences between two sentences. The alignment matrix can further be utilised for interpretation. We then demonstrate that applying de-contextualisation to aligned phrases could help to recognise non-paraphrases of extremely high lexical overlap. Our future work includes exploring other alignment algorithms and more application scenarios for alignment-based sentence encoders.

\section*{Acknowledgement}
We thank all anonymous reviewers for their insightful comments. We would like to further thank Bowen Wang and Wing Yan Li for helpful discussions and proofreading.

\bibliography{acl_latex}
\bibliographystyle{acl_natbib}




\end{document}